\documentclass[journal,transmag]{IEEEtran}
\usepackage{cite}

\usepackage{amsmath,amssymb,amsfonts}

\usepackage{graphicx}
\usepackage{textcomp}
\usepackage{caption}
\usepackage{booktabs}
\usepackage{algorithm}  
\usepackage{algpseudocode}  
\usepackage{amsmath}  

\def\BibTeX{{\rm B\kern-.05em{\sc i\kern-.025em b}\kern-.08em
    T\kern-.1667em\lower.7ex\hbox{E}\kern-.125emX}}

\begin{document}
\title{Compressed Sensing for Photoacoustic Computed Tomography Using an Untrained Neural Network}
\author{Hengrong Lan, Juze Zhang, Changchun Yang, and Fei Gao, \IEEEmembership{Member, IEEE}
\thanks{This paragraph of the first footnote will contain the date on 
which you submitted your paper for review. It will also contain support 
information, including sponsor and financial support acknowledgment. }
\thanks{Hengrong Lan is with the Hybrid Imaging System Laboratory, Shanghai Engineering Research Center of Intelligent Vision and Imaging, School of Information Science and Technology, ShanghaiTech University, Shanghai 201210, China, with Chinese Academy of Sciences, Shanghai Institute of Microsystem and Information Technology, Shanghai 200050, China, and also with University of Chinese Academy of Sciences, Beijing 100049, China (e-mail: lanhr@shanghaitech.edu.cn).}
\thanks{Juze Zhang is with the Hybrid Imaging System Laboratory, Shanghai Engineering Research Center of Intelligent Vision and Imaging, School of Information Science and Technology, ShanghaiTech University, Shanghai 201210, China (e-mail: zhangjz@shanghaitech.edu.cn).}
\thanks{Changchun Yang is with the Hybrid Imaging System Laboratory, Shanghai Engineering Research Center of Intelligent Vision and Imaging, School of Information Science and Technology, ShanghaiTech University, Shanghai 201210, China (e-mail: yangchch@shanghaitech.edu.cn).}
\thanks{Fei Gao is with the Hybrid Imaging System Laboratory, Shanghai Engineering Research Center of Intelligent Vision and Imaging, School of Information Science and Technology, ShanghaiTech University, Shanghai 201210, China (*e-mail: gaofei@shanghaitech.edu.cn).}}

\maketitle

\begin{abstract}
Photoacoustic (PA) computed tomography (PACT) shows great potentials in various preclinical and clinical applications. A great number of measurements are the premise that obtains a high-quality image, which implies a low imaging rate or a high system cost. The artifacts or sidelobes could pollute the image if we decrease the number of measured channels or limit the detected view. In this paper, a novel compressed sensing method for PACT using an untrained neural network is proposed, which decreases half number of the measured channels and recoveries enough details. This method uses a neural network to reconstruct without the requirement for any additional learning based on the deep image prior. The model can reconstruct the image only using a few detections with gradient descent. Our method can cooperate with other existing regularization, and further improve the quality. In addition, we introduce a shape prior to easily converge the model to the image. We verify the feasibility of untrained network based compressed sensing in PA image reconstruction, and compare this method with a conventional method using total variation minimization. The experimental results show that our proposed method outperforms 32.72$\%$ (SSIM) with the traditional compressed sensing method in the same regularization. It could dramatically reduce the requirement for the number of transducers, by sparsely sampling the raw PA data, and improve the quality of PA image significantly.

\end{abstract}


\section{Introduction}
\label{sec:introduction}
\IEEEPARstart{P}{hotoacoustic} imaging (PAI) is a hybrid imaging modality, which originates from the principle of photoacoustic (PA) effect \cite{zhou2016tutorialpat,wang2008review,wang2012pareview,wang2016practicalpai}. The PA signal is induced by a short-pulsed laser light, which propagates in medium and detected by the ultrasonic transducers.  In recent decades, PAI has enabled many interesting imaging applications including hemoglobin and oxygen saturation detection, small animal imaging, and pre-clinical cancer diagnosis \cite{shah2008pacancer,lan2019dualnonPA,gao2015singleSO2,li2017whole-bodypact,lin2018single-breath-hold}. One of the implementations of PAI is photoacoustic computed tomography (PACT), which uses unfocused light to illuminate the tissue, and detects the PA signals by a transducer array.

In PACT, the number of the detector should satisfy the Nyquist sampling theorem. However, increments of the detector will increase the cost of the system. Meanwhile, the transducer could not encircle the field of view (FOV) in some scenes, e.g., the imaging of human carotid. The under-determined setup of the PA image reconstruction is achieved in these cases.

Compressed sensing (CS) has been used to reconstruct the PA image in sparse or limited-view conditions, which could recover the signal/image under the Nyquist sampling rate\cite{provost2008cspa,guo2010compressedpact}.  CS leverages the sparsity of data to reconstruct the PA image based on different optimizations, and uses different priors to solve this inverse problem \cite{provost2008cspa}. For instance, in \cite{provost2008cspa}, the authors firstly compared several sparse representations including Wavelets, Curvelets, Fourier domains, and total variation (TV). Z. Guo et al. adapted CS for PACT reconstruction \cite{guo2010compressedpact}. In this work, CS method was validated in \textit{in-vivo} experiment with Wavelet basis. A modified Curvelet basis was proposed to reconstruct the sparse data in \cite{pan2020pawavelet}. Moreover, many applications of CS are presented to achieve one-shot imaging with a single detector \cite{huynh2019singlecamera,guo2020singleshot}.

Recently, deep learning (DL) is used to reconstruct PA image in sparse view or limited-view conditions \cite{yang2021review,hauptmann2020deepreview,lan2019kigan}, which learns the features of object from numerous data. For instance, Stephan Antholzer et al. used U-Net with residual connection to enhance sparse PA image \cite{antholzer2019sparsepa}.  Stenven Guan et al. proposed a FDU-Net to remove the artifacts of reconstructed image with 10, 15, and 30 detectors \cite{guan2019fdunet}. In \cite{guo2021asnet}, the authors proposed AS-Net to achieve superior results with sparse data. However, these methods need many paired data to pre-train the model. Namely, DL methods require the training of models with enormous amount of data. 
It remains significantly more challenging for PAI since it is hard to acquire a 
large number of data at its infant stage. Moreover, the trained model has difficulty in generalizing for different data.

Inspired by \cite{ulyanov2018dip}, Deep Image Prior (DIP) has been used for 
CS with an untrained convolutional neural networks (CNN) \cite{heckel2020csdip}. 
In this paper, for the first time, we develop and investigate the potential of such approach for 
sparse PACT  reconstruction, which can recover a high quality image with only 50\% 
number of sensors. The additional regularization used in CS can also be used in our method. 
 (TV is demonstrated in this paper.) Furthermore, we introduce a shape 
 prior that penalizes the difference between the output of model with direct reconstruction.
Simulation and experimental data are used to demonstrate this method. The results show that the
proposed method outperforms conventional CS with TV prior. This method bridges the gap between two PA reconstruction schemes: deep learning based CS reconstruction and model-based priors method. Meanwhile, it could be combined with other conventional CS methods.

We list our contributions as follow:

\begin{enumerate}
\item We introduce a CNN model to reconstruct PA image from a few random noise inputs. A CS problem is adapted to an untrained model optimization to approximate the sparse PA signals.  This method has the superiority of not requiring a CNN model trained over the dataset, and addresses the challenge of deep learning methods for the requirement of training dataset.

\item A shape prior is proposed that empirically guides the direction of convergence at the initial iterations. The prior is restricted by the direct reconstructed image. At the first stage, the network initially fits the object; and then, the loss could decrease to fit the artifacts and noise based on empirical DIP. Therefore, the shape prior boosts the model to fit the object before overfitting to artifacts.

\item To implement DIP method in PACT, we decompose the gradient computing process into analytic gradient calculation (the forward  and the adjoint operation) and chained gradient calculation (the CNN model). And then, two parts of gradients will be integrated into back-propagation.

\item We demonstrate this method with conventional regularization (TV regularization in this paper). Simulated and experimental results show that our method outperforms conventional unlearned optimization method with the same regularization. Furthermore, our method embodies a robust and shows generalized performance on different data. Other CS methods are also suitable for integrating into our method, not just the TV regularization.
\end{enumerate}

\section{Background}
\subsection{Photoacoustic Imaging}
In PAI,the initial pressure is excited by a single short laser pulse, which can be expressed as \cite{wang2008review}:

\begin{equation}p_0=\Gamma_0\eta_{th} \mu_a F,\label{1}\end{equation}
where $ \Gamma _0$ is the Gruneisen coefficient, $\eta _{th}$ is the conversion efficiency from light to heat, $\mu _a$ is the optical absorption coefficient, and $F$ is the optical fluence. The pressure propagation in the medium can be described by below equation:

\begin{equation}(\nabla ^2 - \frac{1}{v _s ^2}\frac{\partial ^2}{\partial t^2})p(r,t) 
=-\frac{\beta}{C_P}\frac{\partial H(r,t)}{\partial t},\label{2}\end{equation}
where $p(r,t)$ is the spatiotemporal pressure wave, $v_s$ is the speed of sound, $H$ denotes the heating function, $\beta$ denotes the thermal coefficient of volume expansion, and $C_P$ denotes the specific heat capacity at constant pressure. To compute PA pressure in any heterogeneous medium, we solve this equation with Green function \cite{wang2008review}, and derives the forward model:

\begin{equation}p(\mathbf{r},t)=\frac{1}{4\pi v_s^2}\frac{\partial}{\partial t} \Big[\frac{1}{v_s t}\int d \mathbf{r}'  p_0(\mathbf{r}') \delta(t-\frac{\left|\mathbf{r}-\mathbf{r}'\right|}{v_s}) \Big],\label{3}\end{equation}
where $p_0(\mathbf{r}')$ is the initial pressure at detection position $r'$.We use a linear operator $\mathcal{A}$ indicates the forward procedure from initial pressure distribution $f$ to the PA signals $b$:

\begin{equation} b=\mathcal{A}f+\epsilon,\label{4}\end{equation}
where $\epsilon$ is noise.

The light uniformly illuminates the whole target in PACT, which excites the PA signals simultaneously. The transducer array is used to receive the PA data at different positions. The inversion of Eq.\eqref{3} can be described from $p(\mathbf{r},t)$ to $p_0(\mathbf{r})$ using universal back-projection (UPB) operation \cite{xu2005ubp}: 

\begin{equation}p_0(\mathbf{r})=\frac{1}{\Omega _0}\int _{S_0} \Big[2p(\mathbf{r}_0,t)-\frac{2t\partial p(\mathbf{r}_0,t)}{\partial t} \Big] \frac{cos \theta _0}{\left|\mathbf{r}-\mathbf{r}_0\right|^2} dS_0,\label{5}\end{equation}
where $\theta _0$ is the angle between the vector pointing to the reconstruction point $\mathbf{r}$ and transducer surface $S_0$.

\subsection{Compressed Sensing}
In CS, we use $ \Psi $ as a proper sparsity transform that results in an overdetermined representation, and the sparsity transform can be represented as:

\begin{equation}g=\Psi f,\label{6}\end{equation}
where $f \in \mathbb{R} ^n $ is original data and $g \in \mathbb{R} ^N $ is coefficient on basis $\Psi $.

We can project data $f$ into a series of sensing vector $b$ with noise $\epsilon$, and represents compressive measurements as:

\begin{equation}b=\Phi f+e,\label{7}\end{equation}
we assume it is deterministic noise $\| e \| _2 \le \epsilon$.  We can formulate the original data $f$ obtained by solving the following basis pursuit denoising problem:

\begin{equation} \mathop{\min}_{ f}  \ \| \Psi f \|_1\ \ \mathrm{ s.t. } \|\Phi   f-b \|_2 \le \epsilon.\label{8}\end{equation}

Two conditions should be satisfied if we can successfully recover the ground-truth data $f_0$: 
\begin{itemize}
\item $f _0$ is structured: $\| f_0\|_0 \ll N$;
\item The two basis sets $\Psi$ and $\Phi$ should be incoherent.
\end{itemize}

In the CS theory, we should find a basis $\Psi$ that sparsely represents $f$, and minimize the $l _1$ norm of $\Psi f $ promotes sparsity, and the constraint enforces data consistency. In CS-based PACT, a one-step scheme is described to solve the following minimization problem:

\begin{equation} f^*\longleftarrow \mathop{\arg}\mathop{\min}_{ f}  \frac{1}{2} \|\Phi  \mathcal{A} f-b \|_2 ^2 +\lambda \ \| \Psi f \|_1.\label{9}\end{equation}

\section{Methods}
\subsection{Untrained CNN for CS PACT}
Given the measured PA signals $b$ and the measurement matrix $\mathcal{M}$ ($\mathcal{M}=\Phi \mathcal{A}$), we has $b=\mathcal{M}f+\epsilon$. We aim to recover $f$ from $b$:

\begin{equation} f^*= \mathop{\arg}\mathop{\min}_{ f}  \frac{1}{2} \|  \mathcal{M} f-b \|_2 ^2 + \lambda\mathcal{R}(f),\label{10}\end{equation}
in which $ \|  \mathcal{M} f-b \|_2 ^2$ is the data consistency term, and $\mathcal{R}(f)$ is the regularization term. In CS, the optimal solution of Eq.\eqref{10} depends on $\mathcal{R}(f)$. Namely, we should find a sparse basis as the embedded prior. Some sparse basis has been mentioned before (TV, Wavelets, Curvelets, Fourier), and many papers have studied the use of CNN as NETT regularization for CS-PAT \cite{li2020NETT,antholzer2019NETTCS}. However, a large number of dataset are required to train the model.

Recently, DIP exposed that a generator model is sufficient to capture a great deal of natural images prior without any learning, which can also be used to recover the compressed signal. In our work, we aim to find a set of weights for the output of CNN to fit the reconstructed image, which is applied to the measurement matrix $\mathcal{M}$ by matching the given sparse measured data $b$. To implement that, we should design an over-parametrized CNN decoder model $D$.

Therefore, we initialize the untrained model $\mathcal{M}D(\Theta , z)$ with a fixed random matrix $z$, and solve the non-linear least squares solution:

\begin{equation} \Theta^*= \mathop{\arg}\mathop{\min}_{ \Theta}  \frac{1}{2} \| \mathcal{M} D(\Theta,z)-b \|_2 ^2,\label{11}\end{equation}

Generally, the over-parameterized deep decoder $D$ can fit any image $f^*$, including unstructured noise. Furthermore, an implicit prior can be expressed if we stop the procedure at the correct stage. Namely, further regularization is unnecessary if the procedure of optimization can be early stopped. Also, we can retain the sparse basis as the regularization term like the model-based optimization:

\begin{equation} \Theta^*= \mathop{\arg}\mathop{\min}_{ \Theta}  \frac{1}{2} \| \mathcal{M} D(\Theta,z)-b \|_2 ^2 + \lambda \mathcal{R}(D(\Theta,z)).\label{12}\end{equation}

In this work, a convolutional decoder, $D$, is used as the generator network, and the architecture will be described in the next section.
These CNN models can provide a satisfied prior for natural images in problems such as inpainting and denoising due to their convolutional operations. Therefore, this approach also applies to another differentiable forward operator $\mathcal{A}$, not only PA forward operator.

Note that our method is a learning-free method since it has not the training phase with a number of dataset. Meanwhile, this method leverages an untrained generative decoder to optimize the weights $\Theta$ of model. By using different model, our results further support a hypothesis that network structure, not representation learning, is the key component in image reconstruction.

\begin{figure}[!t]
\centerline{\includegraphics[width=\columnwidth]{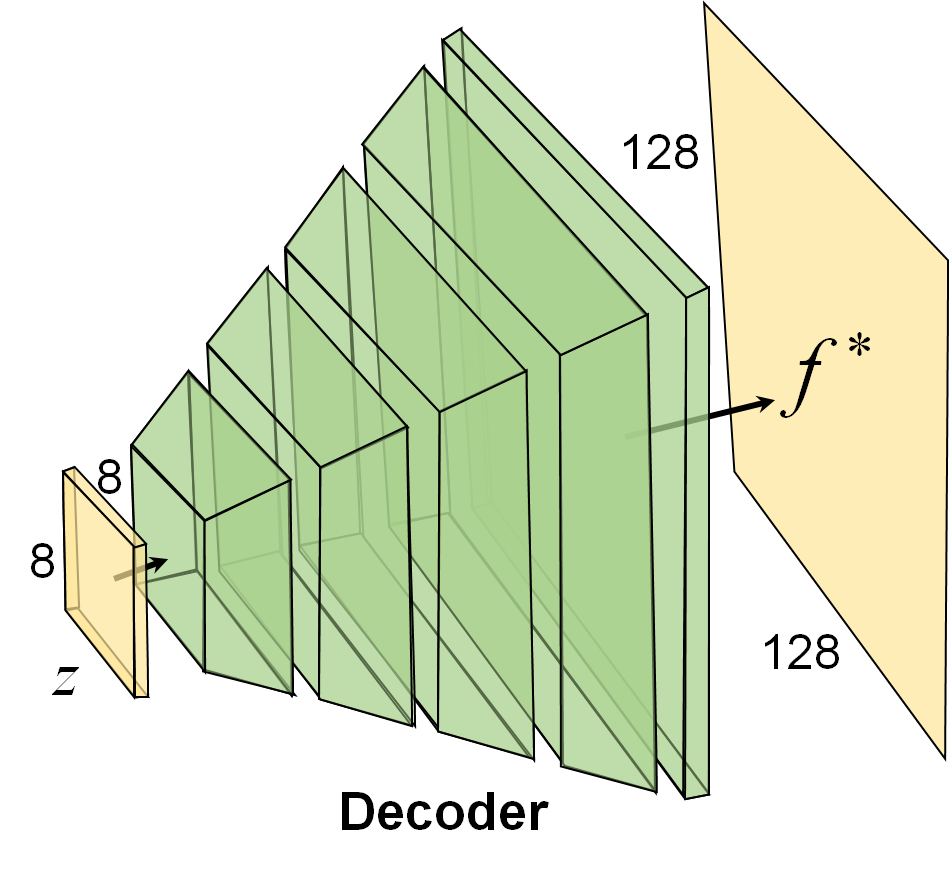}}
\caption{The architecture of proposed decoder $D$, and the detail of each layer has been shown in Eq. \eqref{12}. The input $z$ is a random generated matrix with $8\times8$ size, and output is the fitted image with $128\times 128$ in our work. }
\label{fig1}
\end{figure}

\begin{figure*} [t]
\centerline{\includegraphics[scale=0.38]{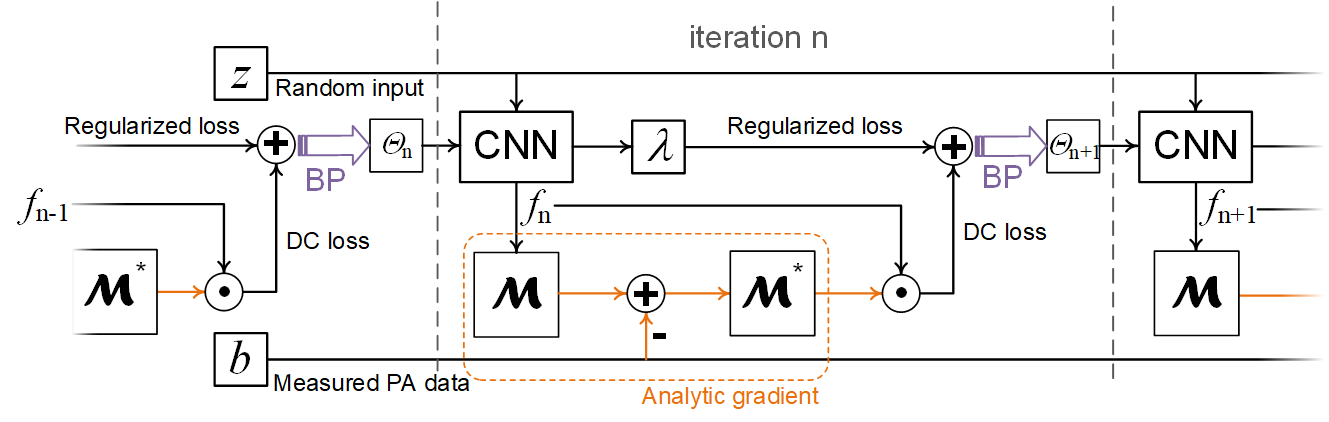}}
\caption{The procedure of optimization of our method. For each iteration, the data consistency loss should be firstly calculated. Then, the other gradients of loss function are computed to back-propagate the gradient. The CNN indicates decoder model $D$ in this paper, $\mathcal{M}$ contains $\mathcal{A}$ and $\Phi$. BP: back-propagation; DC: data consistency. }
\label{fig2}
\end{figure*}

\subsection{Network Architecture}
To demonstrate our method, we introduce a CNN that generates an image through convolutions and non-linear operations. Given a random fixed input $z$, we use a decoder $D$ to generate the final PA image. For $l$ layers' decoder(in our work, $l$=5), the model can be defined as:

\begin{equation} D(z)=  {\rm ReLU}({\rm BN} ({\rm Up}_l (B_l (z)) \ast \kappa _{3\times 3}))  \ast \kappa _{1\times 1} \label{13}\end{equation}
where

\begin{equation}
\begin{split}
B_i = {\rm ReLU}({\rm BN} & ({\rm ReLU}({\rm BN} (B _{\emph{i}-1} \ast \kappa ^1 _i))) \ast \kappa ^2_i) \\
& i=1,...,l, B_0=z.
\label{14}
\end{split}
\end{equation}

Herein, ReLU (Rectified Linear Unit) is an activation function, BN means the batch normalization operation. $\kappa$ is the convolutional kernel, and $\kappa _i$ is $3 \times 3$ size. Note that $B _i$ contains the coefficient of the convolutional layer.

As mentioned above, a five-layer decoder $D$ is used to fit the initial pressure $f^*$, and $D$ is implemented here by Eq. \eqref{13} as shown in Fig. \ref{fig1}. This architecture is a U-Net \cite{ronneberger2015unet} without the encoder and skipped connection.
In this paper, the input data $z$ is a Gaussian random matrix with $8\times8$ size, which should be fixed in the  procedure  of  optimization. A decoder model generates an image with $128\times 128$ size through five convolutional layers and de-convolution. For each layer, double combinations of convolution with $3\times 3$ kernel size, BN, and ReLU are used in series and followed by a de-convolution to up-sample the feature map.

The output image, multiplied by the matrix $\mathcal{M}$, should be restricted by measured raw PA data $b$. Namely, we can directly optimize this model by minimizing the data consistency (DC) loss function:

\begin{equation}  \mathcal{L} _{DC}(\Theta) = \frac{1}{2} \| \mathcal{M} D(\Theta,z)-b \|_2 ^2,\label{15}\end{equation}

\subsection{Shape Prior}
In CS-based PACT, different sparse basis is used. For instance, TV regularization can enforce smoothness as $\mathcal{R}(f)$ in CS theory. Furthermore, the $l_1$ norm of TV regularization can suppress the small coefficients, whose  solution can be sparse. The TV regularization can be described that:

\begin{equation}  {\rm TV}(f) = \| \nabla f \|_1 ,\label{16}\end{equation}

In our method, TV regularization also penalizes the output of the decoder. Therefore, an additional TV term can be contained to restrain the deep generative model, i.e., ${\rm TV}(D(\Theta,z))$. Furthermore, this scheme has the advantage that we do not consider the differentiability of the regularization term ($l_1$ norm), since we optimize the loss function by back-propagation and gradient descent (GD). Now, we rewrite the loss function based on Eq.\eqref{12} and Eq.\eqref{16}:

\begin{equation}
\mathcal{L}(\Theta) =   \frac{1}{2} \| \mathcal{M} D(\Theta,z)-b \|_2 ^2  
+ \lambda {\rm TV}(D(\Theta,z)).\label{17}
\end{equation}

We can iterate this procedure and update the weight with GD. The solo data could cause the stochastic direction of gradient. We further propose a shape prior to improve the performance and create a robust, efficient objective function. Considering that a direct texture of the target could provide a guided optimization at the beginning phase, we calculate the error between rough image and output of $D$ as shape prior (SP). 

In our work, shape prior is proposed to penalize the output of the model, and the rough texture is created by sparse conventional reconstruction. Therefore, we estimate the prior with the decoder model by minimizing the least squares loss $ \|D(\Theta,z)-f_d\|_2 ^2$, and $f_d$ comes from the conventional reconstruction. We combine this prior with the loss function in Eq.\eqref{17}. Finally, we optimize the weights of the Decoder model by minimizing the final loss function as follow :

\begin{equation}
\begin{split}
&\mathcal{L}_{final}(\Theta) =   \frac{1}{2} \| \mathcal{M} D(\Theta,z)-b \|_2 ^2   \\
&+ \lambda _1 {\rm TV} (D(\Theta,z)) +\lambda _2 \frac{1}{2} \|D(\Theta,z)-f_d\|_2 ^2,\label{18}
\end{split}
\end{equation}
$\lambda _1$ and $\lambda _2$ are hyperparameters, $f_d$ is a direct reconstruction from UBP \cite{xu2005ubp} or time-reversal \cite{treeby2010patr}. Note that, for PACT, this function cannot optimize the model directly since the gradient of data consistency term cannot be tracked completely with the chain rule. We introduce a decomposed gradient descent to resolve this problem, which will be described in the next section.

Moreover, our main result shows that the estimate $\Theta$, obtained by running gradient 
descent on the loss until convergence, yields an output $D(\Theta ,z)$ which is close
to $f^*$, i.e., $D(\Theta, z) \approx f ^*$.

\subsection{Implementation}
Generally, the proximal gradient method is used to solve TV minimization in Eq.\eqref{10}:

\begin{equation}
 f ^{n+1} =   {\rm prox} _{\mathcal{R},\alpha} \Big(f^n -\alpha \mathcal{A}^* \Phi^T (\Phi \mathcal{A}f^n -b)\Big).\label{19}
\end{equation}
The forward and adjoint operator $\mathcal{A}$ and $\mathcal{A} ^*$ are used to compute the gradient of the data consistency, which has been implemented in the MATLAB toolbox k-Wave \cite{treeby2010kwave}. For DL model, we do not consider the analytic gradient of loss function since GD and back-propagation are used to update the weight at each iteration. However, in our work, the data consistency contains $\mathcal{A}$ and $D$. Namely, $\mathcal{A}$ and $\mathcal{A} ^*$ cannot be back-propagated, and the gradient of $D$ cannot be calculated directly.

\begin{figure}[ht]
\centerline{\includegraphics[width=\columnwidth]{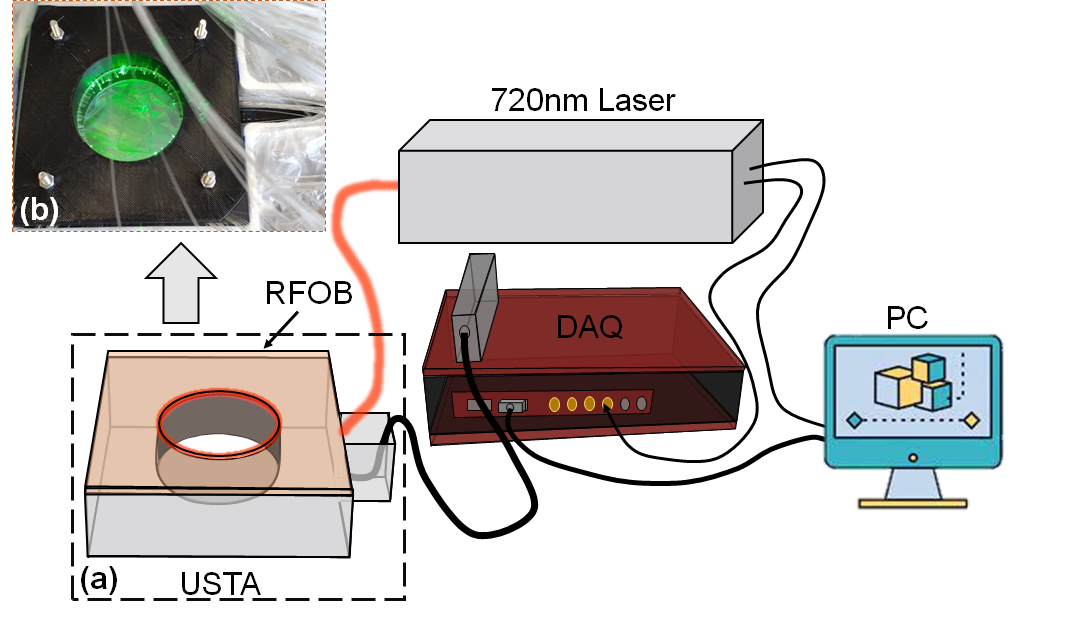}}
\caption{ Schematics of the PACT system. The black box region (a) has a detailed photograph in (b). RFOB: ring-shaped fiber optics bundle; USTA: ultrasonic transducer array; DAQ: data acquisition system; PC: personal computer.}
\label{fig3}
\end{figure}

On the other hand, the forward operator can be discretized and written as a matrix, which is limited by computing resources. The matrix-related gradient cannot be tracked since the size of the matrix is large. Therefore, no matter whether we use the function or matrix, we cannot directly update the weights by back-propagation. The key solution is to decompose the gradient calculation of forward operation and DL model.

To decouple the data consistency term, we compute the gradient of Eq.\eqref{15} for $D$:

\begin{equation}  \frac{\partial \mathcal{L} _{DC}(\Theta)}{\partial D} = \mathcal{M}^*  \big( \mathcal{M}D(\Theta ,z) -b \big),\label{20}\end{equation}
which can be calculated by k-Wave based on the output of $D$. We rewrite the derivative of Eq.\eqref{15} for $\Theta$ based on chain rule:

\begin{equation}  \frac{\partial \mathcal{L} _{DC}(\Theta)}{\partial \Theta} = \frac{\partial \mathcal{L}(\Theta)}{\partial D} \frac{\partial D}{\partial \Theta}. \label{21}\end{equation}
For $\partial D / \partial \Theta$, the weights automatically optimize based on the chain of the gradient. Therefore, the gradient is decomposed into two terms, the first term can be computed by Eq.\eqref{20}, and the second term can be updated by back-propagation. To transfer the gradient of first term, we multiply these two terms and update the weight of DL model by back-propagation. Thus, the gradient of the data consistency term can be transferred to $\Theta$. The procedure of this optimization can be described in Algorithm 1. We can calculate this loss and optimize the weights by back-propagation. We decompose the procedure as back-propagation and analytic gradient descent. For each iteration, the data consistency loss (lines 3 in Algorithm 1) is calculated based on the output of $D$ since $\partial D / \partial \Theta$ can be regarded as a constant for $\Theta$. Next, this loss needs to be combined with other regularization to form the final loss. Finally, the back-propagation is used to update the weights. In Fig.\ref{fig2}, we further illustrate the pipeline of this optimization, and CNN is our decoder model $D$ in this paper.

\begin{algorithm}[h]  
  \caption{The untrained CNN reconstructs the CS-PACT image}  
  \begin{algorithmic}[1]  
    \Require  
      The measurement matrix, $\mathcal{M} (\mathcal{M}=\mathcal{A}\Phi)$,  
      the direct reconstructed image, $f _d$,  
      the measured PA data, $b$,  
      the hyperparameters, $\lambda _1$ and $\lambda _2$. 
    \Ensure  
      The final reconstructed image, $f^*$.
    \State Initialize a random input of model $z$, and the weights of model $\Theta _0$; 

    \For{$i=1$ to $n$}  
      \State $\mathcal{L} _{DC}(\Theta _{i-1}) =\big[\mathcal{M}^*  \big( \mathcal{M}D(\Theta _{i-1} ,z) -b \big) \big] D(\Theta _{i-1},z);$ 
      \State $\mathcal{L} _{TV}(\Theta _{i-1}) ={\rm TV}(D(\Theta _{i-1},z));$
      \State $\mathcal{L} _{SP}(\Theta _{i-1}) =0.5 \times \|D(\Theta _{i-1},z)-f_d\|_2 ^2;$
      \State $\mathcal{L} _{final}(\Theta _{i-1}) =\mathcal{L} _{DC}(\Theta _{i-1}) +\lambda_1 \mathcal{L} _{TV}(\Theta _{i-1}) +\lambda_2 \mathcal{L} _{SP}(\Theta _{i-1});$
      
      \State $\Theta _i \leftarrow {\rm BP} \big( \mathcal{L} _{final}(\Theta _{i-1}) \big)$; $//$ Update $\Theta$ using back-propagation;
      
    \EndFor

    \State $f ^* =D(\Theta _n,z)$. 
 
    \label{algo}  
  \end{algorithmic}  
\end{algorithm}  

In this paper, the optimizer of all experiments is RMSRrop, and the size of output image is 128 $\times$ 128 . We implement this algorithm including $\mathcal{M}$, $\mathcal{M}^*$, and the framework $D (\Theta,z)$ by MATLAB. The initial step rate is 0.001. All methods are implemented on a 64-bit operating system with an Intel Core i7-6700 CPU and an NVIDIA GTX 1080 Ti GPU.

\begin{figure}[ht]
\centerline{\includegraphics[width=\columnwidth]{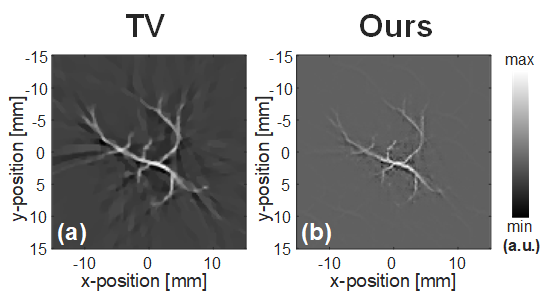}}
\caption{ The synthetic vessel result (50\% sub-sampling rate), (a) the conventional TV method; (b) our approach with TV prior.}
\label{fig4}
\end{figure}

\begin{figure*} [tb]
\centerline{\includegraphics[scale=0.4]{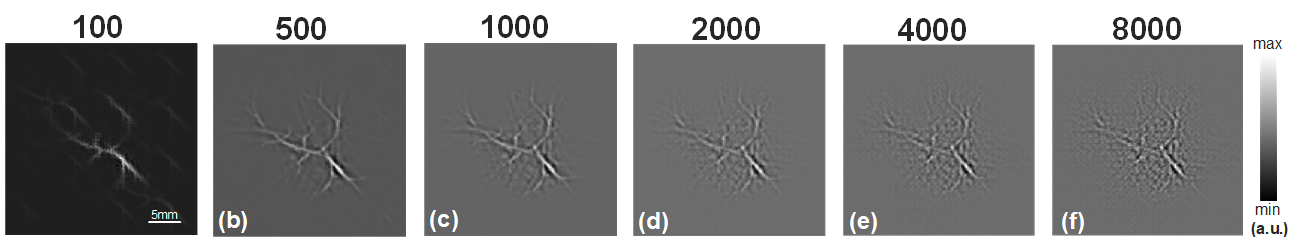}}
\caption{The generated PA images at different iterations, without any regularization. }
\label{fig5}
\end{figure*}

\section{Experiments}
To experimentally validate our approach, simulation data and \textit{in-vivo} data are used. Furthermore, we compare our method with other methods. To validate CS based PACT, the data is 128 channels, and a 50\% sub-sampling matrix is used to sub-sample the channel number. 

Conventional method with TV norm is compared with our method, which leverages Eq.\eqref{19} to solve the objective function. Since our method is unlearned, we only compare it to other unlearned methods. Meanwhile, some ablation studies are demonstrated, e.g., different regularization. Moreover, through experiments, we illustrate the effects of the priors and determine the appropriate number of iterations.

\subsection{Synthetic Setup}
For the synthetic data, we use k-Wave to generate the data. 128 elements circular ultrasound (US) array receives the PA signals with 14.5mm radius. The pixel number of the initial pressure map is 380 $\times$ 380, and the total grid size is 30mm $\times$ 30mm. The sampling rate of PA signal is 40 MSa/s, and noise is added to signal with 40 dB SNR. The center frequency of the US transducer is set as 2.5 MHz with 80\% bandwidth, and the speed of sound is 1500 m/s.  The reconstructed region is 30 mm $\times$ 30 mm with 128 $\times$ 128 pixels.

\subsection{In-vivo Data}
Moreover, we also compare our method with the conventional method on the \textit{in-vivo} data, which contains the brain of mice and the cross-section of the human finger. All data is acquired from a self-built PACT system in Fig. \ref{fig3}. The transducer array is a customized 128-elements full-view circular with 30mm radius (2.5 MHz, Doppler Inc.), which is placed in a 3-D printed water tank. The laser source is a pulsed laser (720 nm wavelength, 10 Hz repetition rate), which illuminates the object by a fiber optic bundle as Fig. \ref{fig3} shows. The data sampling rate of data acquisition system is 40 MSa/s. The region of image reconstruction is 30 mm $\times$ 30 mm with 128 $\times$ 128 pixels.

\section{Results}
\label{sec:guidelines}

\subsection{Synthetic Results}
We firstly validate our method on the synthetic data, $\lambda _1$ and $\lambda _2$ are 0.006 and 0.05 respectively. In Fig.\ref{fig4}, we show the results of TV method and our method. Specially, our method is minimized by Eq.\eqref{18}, which refers to this function by default in this paper. This holds by simply running TV method until convergence (300 iterations). Note that, for all experiments, the number of iteration is 700. Due to the reduction in the number of channels, the background of conventional result is polluted, which causes a poor contrast compared with Fig.\ref{fig4}. For our approach, most structures of the object are reconstructed well with few artifact. It shows that the decoder $D$ fits the object at the initial phase, and the artifacts are reconstructed after continuously optimizing. We can compute the Structural Similarity (SSIM) of these results to quantitatively compare the performance. The SSIM values of Fig.\ref{fig4} are 0.6312 and 0.8377 respectively, which also indicate our method outperforms the conventional method over 32\%.

Furthermore, we should validate the effects of different priors and the appropriate iteration times. A series of comparison experiments are set up.
\subsubsection{Iteration times}
{
We use an untrained model $D$ to compare the performance of different numbers of iterations without any regularization. Different iteration times have been validated from 100 to 8000 as Fig.\ref{fig5} shows. As the number of iterations increases, the main object is reconstructed firstly (from 1 to 500), and the best reconstruction is achieved between 500 and 1000. And then, the reconstruction result starts to blur (after 1000) since the artifacts near object are appearing. We should further quantitatively evaluate these results. 

Three metrics are used to quantify the performance of different results, which are SSIM, Peak Signal to Noise Ratio (PSNR), and Signal to Noise Ratio (SNR). TABLE \ref{tab1} demonstrates the quantitative results of these different iterations. The quantitative results show that the reconstruction quality first increases and then decreases with the number of iterations increasing. Namely, the model fits the correct object at the beginning, and the best quality is 500 iterations in TABLE \ref{tab1}. Therefore, for PACT, the best iterative times could be selected from 500 to 1000. After comparing different data, we determined to use 700 iterations for all experiments without unnecessary artifacts.
}

\begin{table}
\caption{The quantitative results of different iteration}
{
\centering
\begin{tabular}{@{}ccccccc@{}}
\toprule
Iterations & 100     & 500     & 1000    & 2000    & 4000    & 8000    \\ \midrule
SSIM       & 0.7418  & 0.8236  & 0.7958  & 0.7591  & 0.7130  & 0.6405  \\
PSNR       & 21.2158 & 21.3040 & 21.1562 & 20.8575 & 20.6561 & 20.5045 \\
SNR        & 1.7469  & 1.8350  & 1.6873  & 1.3886  & 1.1872  & 1.0356  \\ \bottomrule
\end{tabular}
}
\label{tab1}
\end{table}

\begin{figure}[htb]
\centerline{\includegraphics[width=\columnwidth]{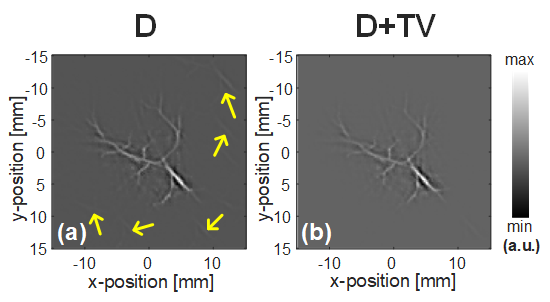}}
\caption{ Ablation studies results, (a) the vessel result only using network $D(\Theta,z)$; (b) the vessel result using network $D(\Theta,z)$ and TV prior (without SP). The yellow arrows indicate some noises in background.}
\label{fig6}
\end{figure}

\subsubsection{Ablation study}
{
For DIP, the DL model can express the implicit prior generally, thus the additional prior term. In this section, we evaluate the effects of two priors (TV and SP) as ablation studies. The synthetic vessel results have been shown in Fig.\ref{fig6}. The two results show similar reconstructions from Fig.\ref{fig6}. The background of $D$'s result has some noises as the yellow arrows indicated in Fig.\ref{fig6} (a). By contrast, Fig.\ref{fig4} has a purer background and higher contrast compared with Fig.\ref{fig6}.

We list the quantitative results of Fig.\ref{fig4} and Fig.\ref{fig6} in TABLE \ref{tab2}. The first three columns of the table indicate that each of the two priors items has improved the reconstructed quality. Although the conventional sparse basis can be surpassed only using a deep model, different regularization can further boost the robustness and efficiency of this method. Compared with the decoder $D$ without regularization terms, the decoder $D$ with regularization performs better in terms of noise suppression, i.e., higher SNR (3.1046 and 1.6595). Similarly, the results of the quantitative comparisons also reflect the same performance. These results further show that effective priors can improve the performance of untrained CNN.
}

\begin{table}[hp]
\caption{The quantitative results of ablation studies}
\centerline{
\begin{tabular}{@{}ccccccc@{}}
\toprule
           & $D$     & $D$+TV  & $D$+TV+SP          & TV      \\ \midrule
SSIM       & 0.8136  & 0.8290  & \textbf{0.8377}    & 0.6312  \\
PSNR       & 21.1285 & 21.7119 & \textbf{22.5736}   & 18.1459 \\
SNR        & 1.6595  & 1.4430  & \textbf{3.1046}    & -1.323  \\ \bottomrule
\end{tabular}
}
\label{tab2}
\end{table}

\subsection{\textit{In-vivo} Results}
\begin{figure}[htb]
\centerline{\includegraphics[width=\columnwidth]{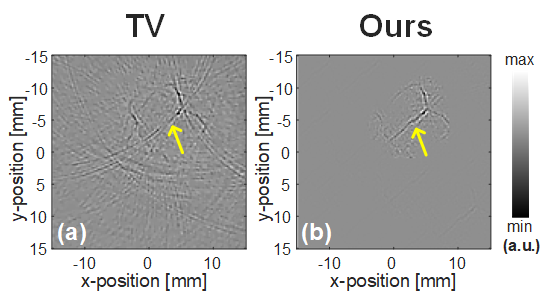}}
\caption{ The \textit{in-vivo} mice brain result (50\% sub-sampling rate), (a) the conventional TV method ; (b) our approach with TV prior.}
\label{fig7}
\end{figure}

In addition, we demonstrate our method on \textit{in-vivo} data, $\lambda _1$ and $\lambda _2$ are 0.005 and 0.1, respectively. Firstly, a real mice brain data is validated. We also compare TV method with our method. Fig.\ref{fig7} shows the brain imaging result, where the TV obtains the result with 300 iterations. Obviously, the conventional method cannot suppress the artifacts only using 64 channels data from Fig.\ref{fig7} (a). The untrained CNN model method shows a superior result, purer background creates a higher contrast in Fig.\ref{fig7} (b). From the yellow arrow in Fig.\ref{fig7}, it shows that the vessel in the sulcus has a clear shape compared with TV result, which contains a few artifacts in Fig.\ref{fig7} (a).

\begin{figure}[htb]
\centerline{\includegraphics[width=\columnwidth]{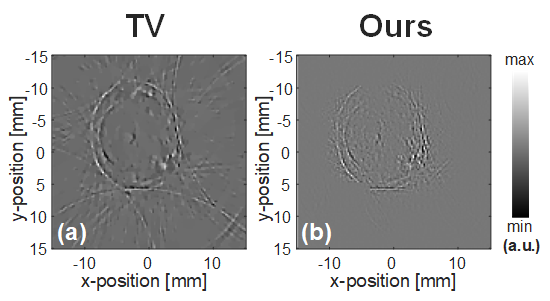}}
\caption{ The \textit{in-vivo} cross section of finger result (50\% sub-sampling rate), (a) the conventional TV method; (b) our approach with TV prior.}
\label{fig8}
\end{figure}

We further use a cross-section of human finger to demonstrate different methods. We also compare these two different methods in Fig.\ref{fig8}. Similarly, residual artifacts are retained in the result of conventional method as shown in Fig.\ref{fig8} (a). The 50\% sub-sampling rate, i.e., 64 channels, causes a blurry result that the object is disturbed by the artifacts. For our result, Fig. \ref{fig8} (b) eliminates most of the obvious artifacts compared with Fig. \ref{fig8} (a). Note that the artifacts near the objects are also beginning to be reconstructed from  Fig.\ref{fig8} (b). However, for the SNR and contrast, our method still completely outperforms the conventional method. In addition, these results further verify the characteristics of this method, which will first fit the target and then fit the noise and the artifacts. It implies this method can fit any signal with appropriately stopping.

\section{Conclusion}
In this paper, we introduce an untrained CNN model to reconstruct a sparse PACT image, which outperforms unlearned methods without plenty of data. In addition, a direct reconstructed image is used to penalize the output of DL model. This prior improves the reconstruction error and efficiency. We further demonstrate how to implement this method on PACT, which further decomposes the analytic gradient and chained gradient in data consistency term. The experimental results show our approach outperforms the conventional CS method with the same sparse basis. Note that DIP method can fit any signal given an over-parameterized model in empirical. This method provides insight for CS based PACT, and explores more solid works combined with other conventional CS methods. Meanwhile, we will compare another sparse basis in future works.



\bibliographystyle{IEEEtran}
\bibliography{ref.bib}

\end{document}